\def\paperID{0000}
\def\confName{CVPR}
\def\confYear{2024}

\def\authorBlock{
Jianqiang Ren,~~Chao He,~~Lin Liu,~~Jiahao Chen,~~Yutong Wang,~~Yafei Song,~~Jianfang Li,\\~~Tangli Xue,~~Siqi Hu,~~Tao Chen,~~Kunkun Zheng,~~Jianjing Xiang,~~Liefeng Bo \\ 
   Institute for Intelligent Computing,~~Alibaba Group\\
   
    {\tt\small \{jianqiang.rjq, yichao.hc, lorrain.ll, peter.cjh, 
 yutong.yutongwang, huaizhang.syf,  wuhui.ljf, } \\
 {\tt\small xuetangli.xtl, husiqi.hsq, ct253279, kunkun.zkk, jianjing.xjj, liefeng.bo\}@alibaba-inc.com}
}

\newif\ifreview 
\newif\ifarxiv \newcommand{\arxiv}{\arxivtrue}
\newif\ifcamera 
\newif\ifrebuttal 

\arxiv  %\review % \review OR \arxiv OR \cameraready

\pdfoutput=1
\documentclass[10pt,twocolumn,letterpaper]{article}
\ifreview \usepackage[review]{cvpr} \fi
\ifarxiv \usepackage[pagenumbers]{cvpr} \fi
\ifrebuttal \usepackage[rebuttal]{cvpr} \fi
\ifcamera \usepackage{cvpr} \fi

\usepackage{graphicx}	
\usepackage{amsmath}	
\usepackage{amssymb}	
\usepackage{booktabs}
\usepackage{times}
\usepackage{microtype}
\usepackage{epsfig}
\usepackage[table,xcdraw,dvipsnames]{xcolor}
\usepackage{caption}
\usepackage{float}
\usepackage{placeins}
\usepackage{color, colortbl}
\usepackage{stfloats}
\usepackage{enumitem}
\usepackage{tabularx}
\usepackage{xstring}
\usepackage{multirow}
\usepackage{xspace}
\usepackage{url}
\usepackage{subcaption}
\usepackage{xcolor}
\usepackage[hang,flushmargin]{footmisc}

\ifcamera \usepackage[accsupp]{axessibility} \fi

% ...

\ifarxiv  \fi

\newcommand{\R}[1]{{%
    \textbf{%
        \ifstrequal{#1}{1}{\textcolor{red}{R#1}}{%
        \ifstrequal{#1}{2}{\textcolor{blue}{R#1}}{%
        \ifstrequal{#1}{3}{\textcolor{magenta}{R#1}}{%
        \ifstrequal{#1}{4}{\textcolor{teal}{R#1}}{%
                           \textcolor{cyan}{R#1}%
        }}}}%
    }%
}}
\usepackage{xr-hyper}

\makeatletter
\newcommand*{\addFileDependency}[1]{
  \typeout{(#1)}
  \@addtofilelist{#1}
  \IfFileExists{#1}{}{\typeout{No file #1.}}
}

\makeatother
\newcommand*{\myexternaldocument}[1]{
    \externaldocument{#1}
    \addFileDependency{#1.tex}
    \addFileDependency{#1.aux}
}

\definecolor{cvprblue}{rgb}{0.21,0.49,0.74}
\usepackage[pagebackref,breaklinks,colorlinks,citecolor=cvprblue]{hyperref}
\usepackage[capitalize]{cleveref}
\crefname{section}{Sec.}{Secs.}
\crefname{table}{Table}{Tables}
\crefname{figure}{Fig.}{Figs.}

\ifarxiv \crefname{appendix}{App.}{Apps.}
\else \crefname{appendix}{Suppl.}{Suppls.} \fi

\usepackage{indentfirst}
\unless\ifarxiv \myexternaldocument{_supplementary} \fi

\begin{document}
%% TITLE
\title{Make-A-Character: \\ High Quality Text-to-3D Character Generation within Minutes}
\author{\authorBlock}
\maketitle

\begin{abstract}

There is a growing demand for customized and expressive 3D characters with the emergence of AI agents and Metaverse, but creating 3D characters using traditional computer graphics tools is a complex and time-consuming task. To address these challenges, we propose a user-friendly framework named Make-A-Character (Mach) to create lifelike 3D avatars from text descriptions. The framework leverages the power of large language and vision models for textual intention understanding and intermediate image generation, followed by a series of human-oriented visual perception and 3D generation modules. Our system offers an intuitive approach for users to craft controllable, realistic, fully-realized 3D characters that meet their expectations within 2 minutes, while also enabling easy integration with existing CG pipeline for dynamic expressiveness. For more information, please visit the project page at \url{https://human3daigc.github.io/MACH/}.
 
\end{abstract}
\section{Introduction}
\label{sec:intro}
Realistic-looking 3D avatars have been widely utilized in the realms of video games, VR/AR, and film industry. With the rise of the Metaverse and AI agents, the demand for personalized and expressive character creation has surged in fields like virtual meetings, conversational agents, and intelligent customer service. However, for general users, creating a personalized 3D avatar using traditional digital creation tools is a complex and time-consuming task. To lower the barrier to 3D digital human creation, this work unveils an innovative system, named Make-A-Character (Mach), which harnesses the power of large language and vision foundation models to generate detailed and lifelike 3D avatars from simple text descriptions. Our system seamlessly converts textual descriptors into vivid visual avatars, providing users with a simple way to create custom avatars that resonate with their intended personas.

We summarize the properties of our generated 3D characters as follows:

\noindent {\bf Controllable.} Our system empowers users with the ability to customize detailed facial features, including the shape of the face, eyes, the color of the iris, hairstyles and colors, types of eyebrows, mouths, and noses, as well as the addition of wrinkles and freckles. This customization is facilitated by intuitive text prompts, offering a user-friendly interface for personalized character creation.

\noindent {\bf Highly-Realistic.} The characters are generated based on a collected dataset of real human scans. Additionally, their hairs are built as strands rather than meshes. The characters are rendered using PBR (Physically Based Rendering) techniques in Unreal Engine, which is renowned for its high-quality real-time rendering capabilities.

\noindent {\bf Fully-Completed.} Each character we create is a complete model, including eyes, tongue, teeth, a full body, and garments. This holistic approach ensures that our characters are ready for immediate use in a variety of situations without the need for additional modeling.

\noindent {\bf Animatable.} Our characters are equipped with sophisticated skeletal rigs, allowing them to support standard animations. This contributes to their lifelike appearance and enhances their versatility for various dynamic scenarios.

\noindent {\bf Industry-Compatible.} Our method utilizes explicit 3D representation, ensuring seamless integration with standard CG pipelines employed in the game and film industries. 

% In the rest of this report, we first overview the whole system and then illustrate each key component. Some evaluation results are presented to illustrate the system's efficacy in producing diverse, realistic 3D avatars that meet the expectations of users.

%\input{02_related}

\section{Method}
\label{sec:method}

\begin{figure*}[ht]
  \centering
  \resizebox{0.98\linewidth}{!}{
   \includegraphics{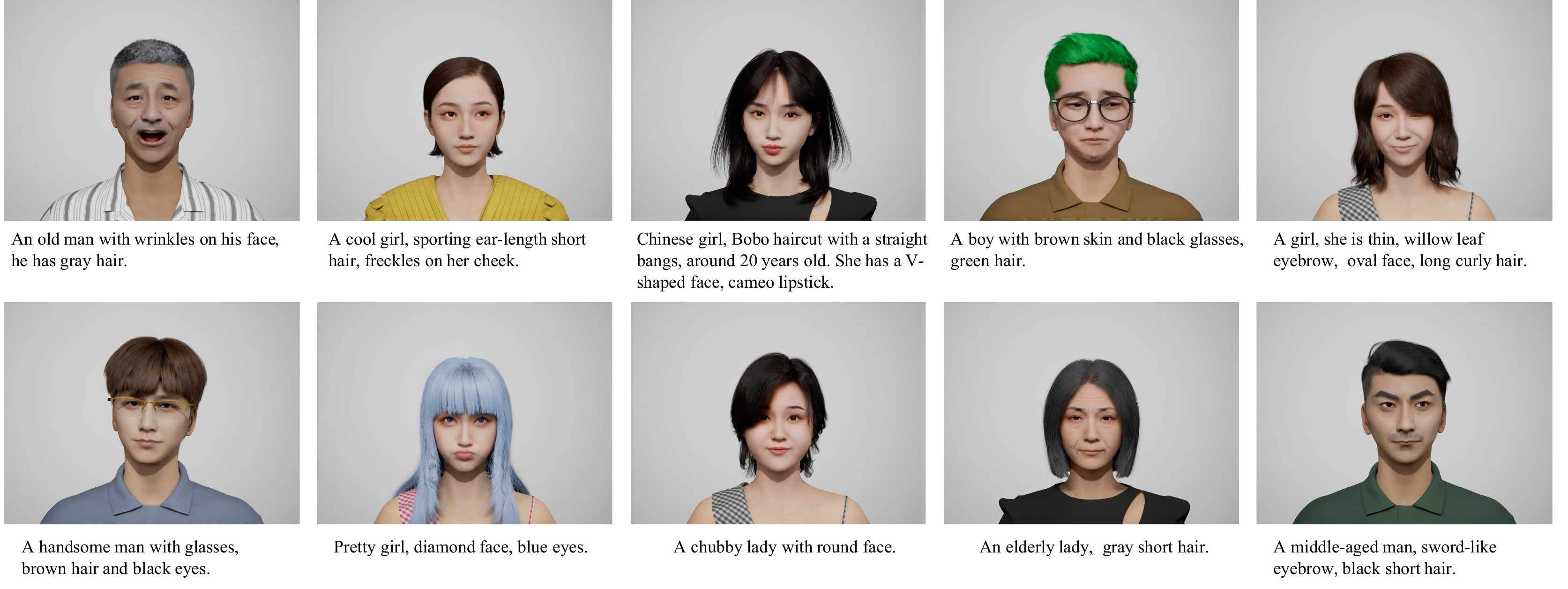} }
  \caption{The Mach system is capable of creating highly detailed and varied 3D avatars from text prompts. We demonstrate the versatility of these characters by showcasing their ability to express dynamic animations through various facial expressions.}
    \vspace{-15pt}
  \label{fig: generation_showcases}
\end{figure*} 

Mach aims to create complete, lifelike, drivable 3D virtual avatars that are compatible with existing CG pipeline and offer flexible styling and animation ability. Therefore, we have opted for an explicit 3D representation(\emph{i.e.}, surface mesh and texture) rather than an implicit approach like NeRF. In terms of geometric base model selection, we conduct research on various models including BFM, FLAME~\cite{2017Learning}, Daz 3D~\cite{Daz3D}, Metahuman~\cite{Metahuman}, and SMPL~\cite{SMPL:2015}. Ultimately, we choose MetaHuman because it includes a complete representation of the face and body, and it offers more nuanced expression animation capabilities, primarily because of its advanced facial rig system~\cite{RigLogic}, which offers a powerful support for the vivid dynamic performance of virtual characters.

\begin{figure*}[ht]
  \centering
  \resizebox{0.98\linewidth}{!}{
   \includegraphics{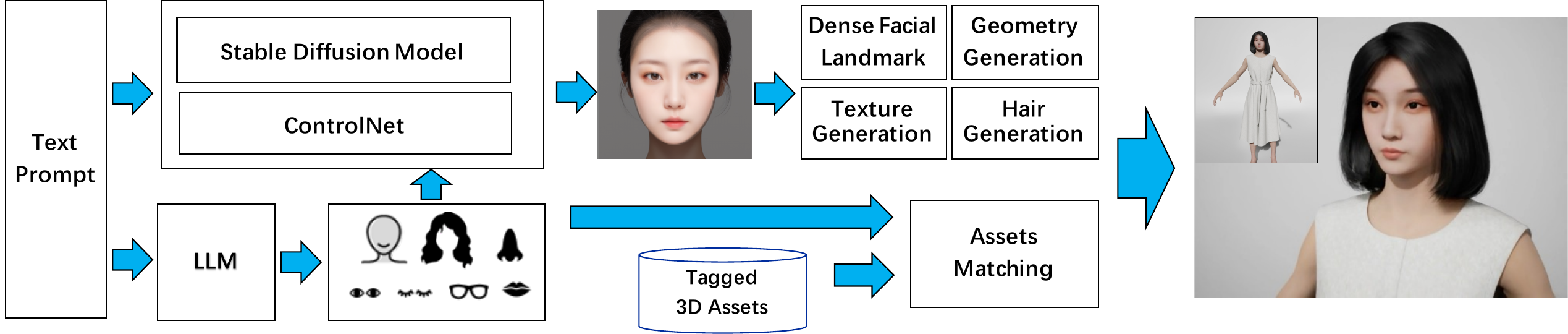} }
  \caption{The framework utilizes the Large Language Model (LLM) to extract various facial attributes(\emph{e.g.}, face shape, eyes shape, mouth shape, hairstyle and color, glasses type). These semantic attributes are then mapped to corresponding visual clues, which in turn guide the generation of reference portrait image using Stable Diffusion along with ControlNet. Through a series of 2D face parsing and 3D generation modules, the mesh and textures of the target face are generated and assembled along with additional matched accessories. The parameterized representation enable easy animation of the generated 3D avatar.}
  \label{fig: architecture}
  % \vspace{-15pt}
\end{figure*}

The architecture of Mach is illustrated in Figure~\ref{fig: architecture}. Given a text prompt, The Large Language Model (LLM) is leveraged for semantic comprehension, enabling the extraction of various facial attributes, such as face shape, eyes shape, mouth shape, hairstyle and color, glasses type. Some of these semantic attributes are then mapped to corresponding visual clues, which serve as fine guidance for generating a reference portrait image using Stable Diffusion~\cite{rombach2022high} and ControlNet~\cite{zhang2023adding}. The reference portrait is guaranteed to be frontal with neutral expression owing to our posture control, which brings great convenience to head geometry and texture generation. We build a conversion mechanism between head mesh and triplane maps, thus we can directly optimize 2D maps instead of resorting to 3DMM methods, which offers flexible vertex-level control. Differentiable rendering and delighting techniques are utilized to extract and refine diffuse texture based on the reference image, and our hair generation module enhances the overall expressiveness by providing strand-level hair synthesis. For other accessories such as garments, glasses, eyelashes, and irises, we match them from the tagged 3D asset library with extracted semantic attributes, and finally assemble them into a complete 3D figure. The duration of the entire process is within 2 minutes. Detailed introductions to each module are provided in the following sections. 
 
\subsection{LLM-Powered Visual Prompt Generation}
\begin{figure}[t]
  \centering
  \resizebox{0.98\linewidth}{!}{
   \includegraphics{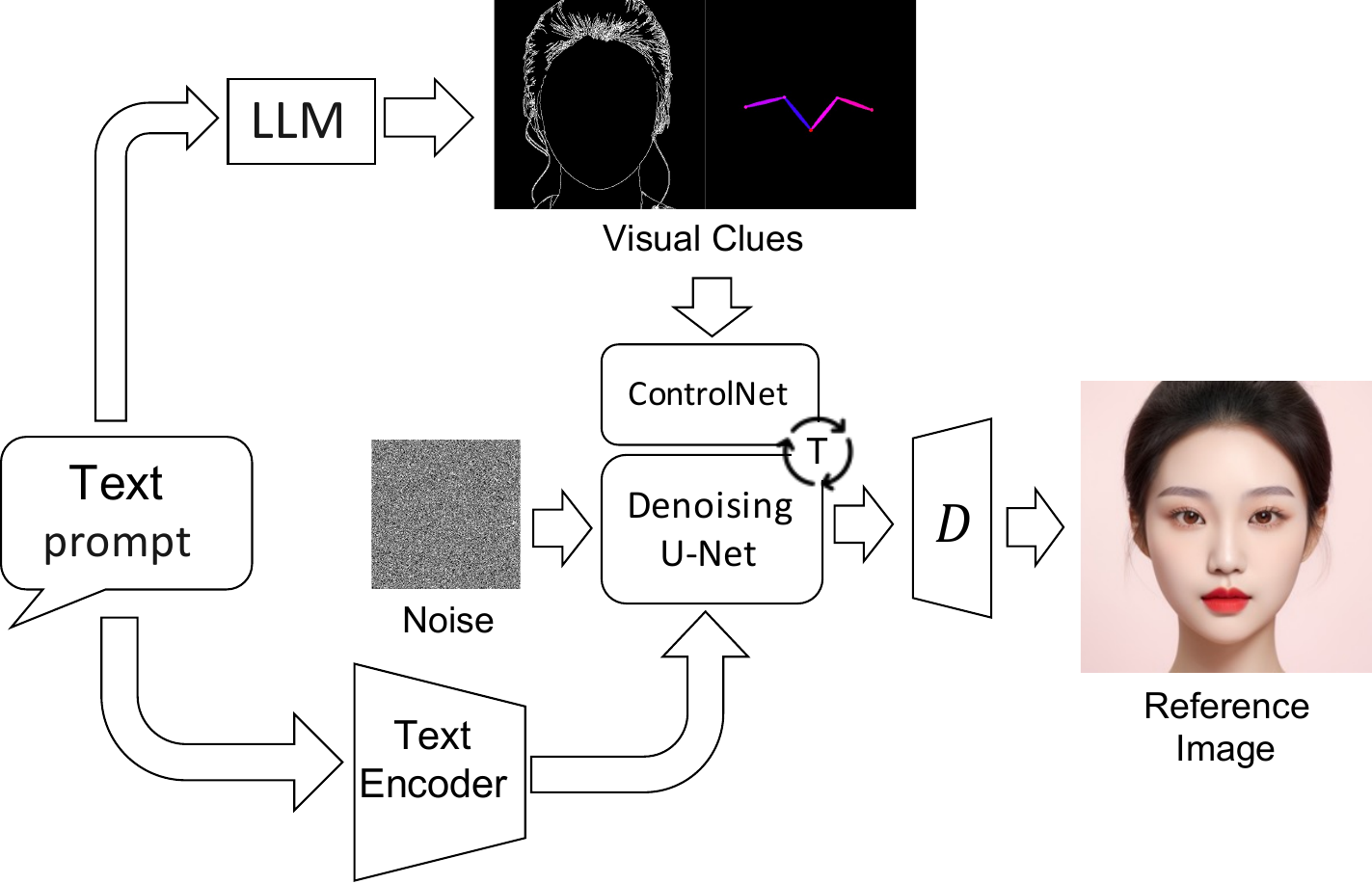} }
%   \vspace{-10pt}
  \caption{We find that generating corresponding detailed facial attributes guided by only text prompt using the Stable Diffusion Model is challenging due to the absence of comprehensive facial annotations and corresponding image pairs. To address this issue, we employ the Large Language Model (LLM) to extract attributes and align them with low-level visual clues, such as posture and edge maps. These visual clues then direct the text-to-image (T2I) generation process via a ControlNet, enhancing the model's ability to accurately render facial details.}
  \label{fig: SD_framework}
  \vspace{-15pt}
\end{figure}

The utilization of large models is illustrated in Figure~\ref{fig: SD_framework}. Due to Stable Diffusion's insensitivity to subtle facial attributes (including face shape, eyebrows, eye shape, nose, mouth, etc.), it fails to provide finer-grained control over these attributes. To address this limitation, we perform facial attributes analysis on the text prompt using Qwen-14B~\cite{qwen} to acquire visual clues related to these attributes, and then apply ControlNet to regulate the fine-grained features of facial components. In the deployment of ControlNet, we integrate Openpose and canny maps to ensure a reasonable distribution of facial features, eventually obtaining reference images that are strongly correlated with the text prompts.

\subsection{Dense Landmark Detection}
Face landmarks refer to identifiable points on face that correspond semantically across different individuals, such as the nose tip and mouth corners. They typically represent the face geometry on 2D image domain and are essential for reconstructing a 3D face from a single image. The traditional 68 or 98 face landmarks in ~\cite{6755925} and ~\cite{wayne2018lab} are considered sparse landmarks,  meaning they cover only a limited area of face and leave regions like the forehead and cheeks without landmarks. The absence of landmarks in these areas such as the forehead makes it challenging to reconstruct the structure of the forehead accurately.

\begin{figure}[t]
  \centering
 
    \begin{subfigure}{0.5\linewidth}
        \includegraphics[width=\linewidth]{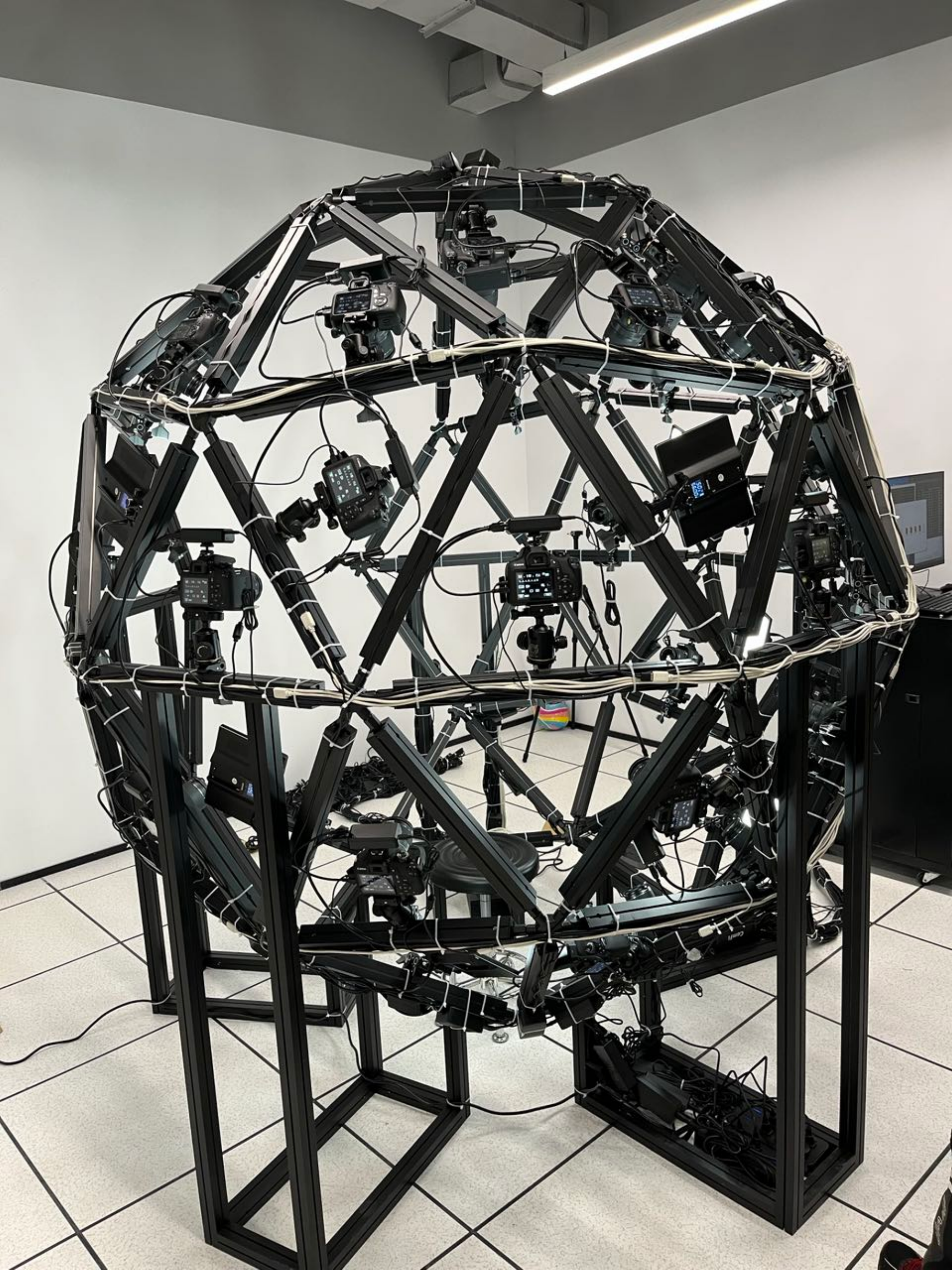}
        \caption{Face Scanner}
    \end{subfigure}\hspace{-1mm}
    \begin{subfigure}{0.5\linewidth}
        \includegraphics[width=\linewidth]{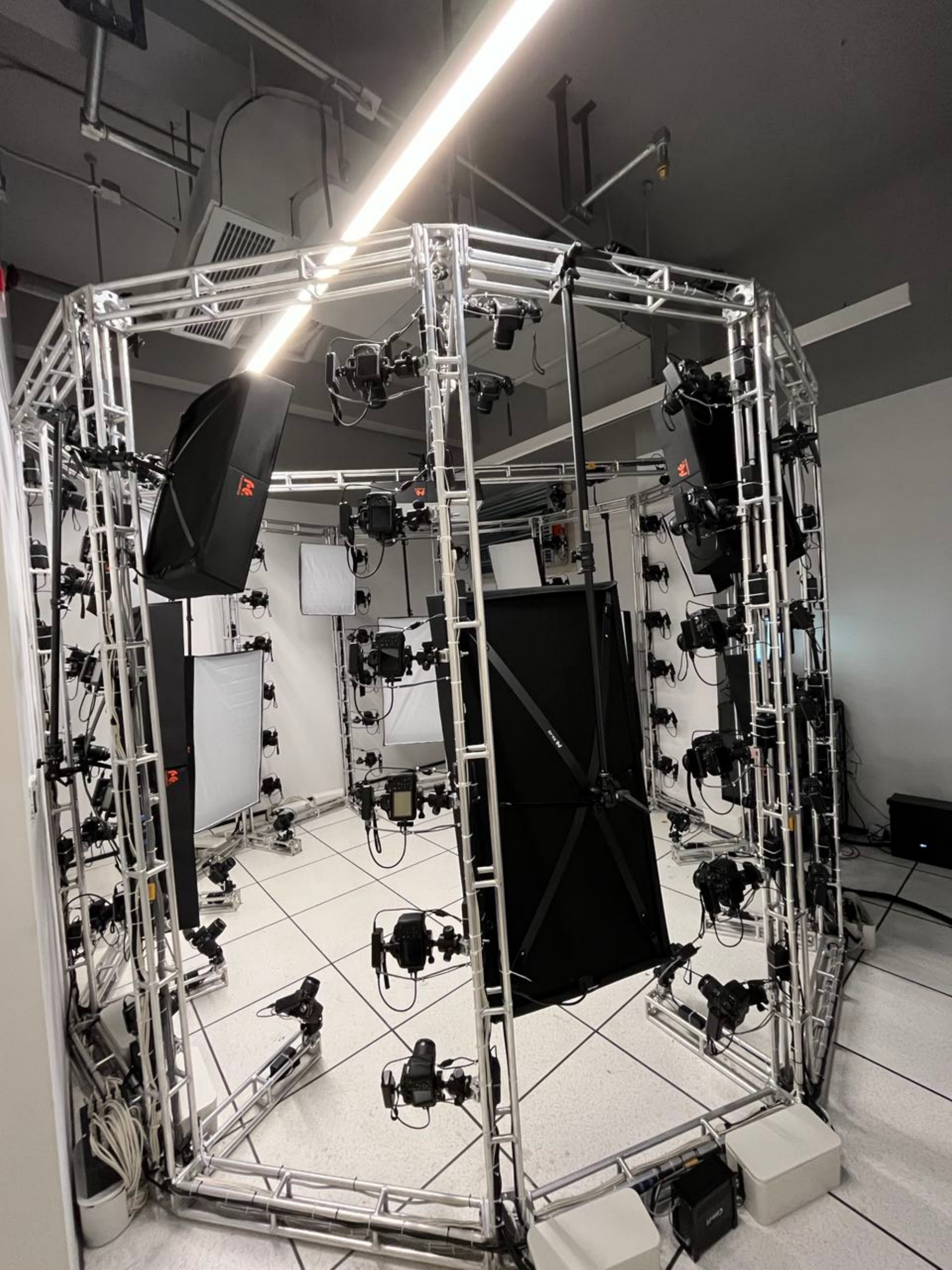}
        \caption{Body Scanner}
    \end{subfigure}\hspace{-1mm}
  \caption{Our multi-view light stage setups are designed for capturing high-resolution 3D scans of heads and bodies.}
  \label{fig: Light_Stage}
  % \vspace{-15pt}
\end{figure} 

To overcome these limitations, previous works utilize supplementary information like image colors~\cite{deng2019accurate, Feng:SIGGRAPH:2021}, optical flow~\cite{10.1145/3272127.3275093}. However, the reliability of such data on face images is often questionable, as it can change drastically with varying lighting conditions and different camera viewing angles. In contrast, we utilize dense facial landmarks as the primary information to reconstruct face and head geometry inspired by ~\cite{wood2022dense}. Since dense facial landmarks cannot be annotated manually, we follow the work ~\cite{wood2022dense} to adopt synthetic images for training data.

We established a multi-view capturing and processing pipeline (Figure~\ref{fig: Light_Stage}) to produce uniformly topological head geometries and facial textures from 1,000 real human scans. Additionally, we utilized a variety of digital assets, including 23 hairs, 45 clothes, 8 hats, and 13 moustaches, to create complete human head models. For generating different facial expressions, we employed the classic set of 52 blendshapes and rendered each model from multiple angles. 
For landmark detection, we employ the stacked hourglass networks~\cite{Newell2016StackedHN} to regress heat maps for each facial landmark. Since the systematic definition of landmarks is different from traditional sparse landmarks,  we conducted comparisons by calculating re-projection errors on the FaceScape~\cite{yang2020facescape} dataset, Specifically, we measured the re-projection errors for 98 landmarks excluding those along the jawline. The results are presented in Table~\ref{table:DenseLMK:error}.

\begin{table}[]
\centering
\begin{tabular}{l|cc}
\toprule
Method    & Is Dense & Re-projection error \\
\midrule
StarLoss~\cite{Zhou_2023_CVPR}  & Not   & 4.00                \\
MediaPipe & Yes   & 5.31                \\
Ours       & Yes   & 3.19           \\
\bottomrule
\end{tabular}\caption{Re-projection errors on FaceScape~\cite{yang2020facescape} dataset.}\label{table:DenseLMK:error}
\end{table}

\begin{figure}[ht]
  \centering
    \begin{subfigure}{0.33\linewidth}
        \includegraphics[width=\linewidth]{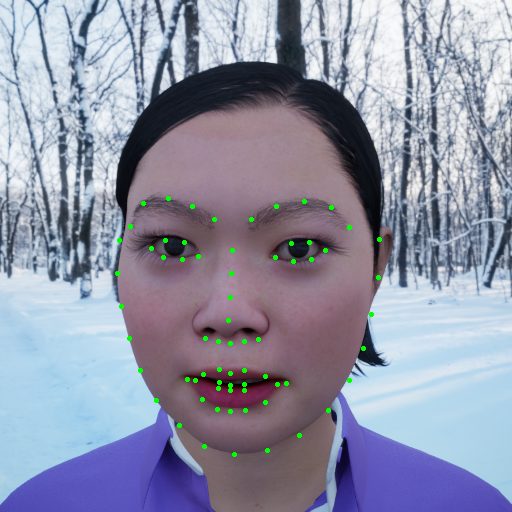}
        \caption{68 Landmarks}
    \end{subfigure}\hspace{-1mm}
    \begin{subfigure}{0.33\linewidth}
        \includegraphics[width=\linewidth]{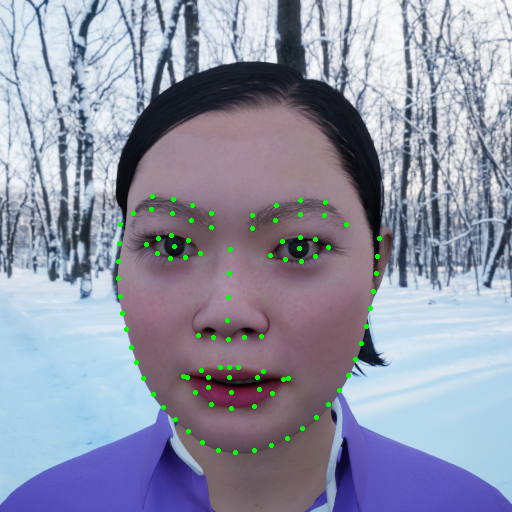}
        \caption{98 Landmarks}
    \end{subfigure}\hspace{-1mm}
    \begin{subfigure}{0.33\linewidth}
        \includegraphics[width=\linewidth]{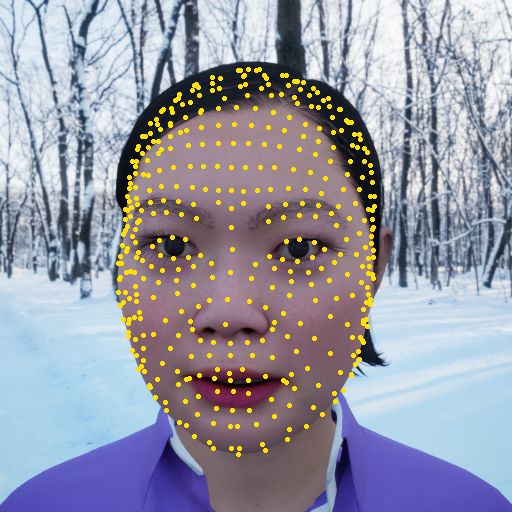}
        \caption{431 Landmarks}
    \end{subfigure}
  \caption{The traditional 68 landmarks~\cite{6755925}, 98 landmarks~\cite{wayne2018lab}, and our 431 landmarks. The traditional landmarks are sparse on face. In contrast, our 431 landmarks are dense landmarks that cover the whole head.}
  \label{fig:DenseLMK}
  % \vspace{-15pt}
\end{figure}

\subsection{Geometry Generation}
Once given the reference portrait image and corresponding dense facial landmarks, we reconstruct the head geometry under the guidance of these landmarks. We firstly establish a conversion mechanism between 3D mesh and 2D maps, this is accomplished by mapping each vertex's position onto three orthogonal planes (\emph{i.e.} Y-Z, X-Z, and X-Y planes) in accordance with its UV coordinates, thus we can represent the 3D mesh with a 3-channels image, referred to as triplane. This representation enhances the potential for geometry generation through a 2D diffusion model and facilitates the local smoothness constraints. 
For each 2D landmarks $p_k$, we predefine its vertex correspondence $V_k$ on base mesh, and introduce the landmark projection loss as:
\begin{equation} \label{eq: landmark_loss}
\mathcal{L}_{lmk} =\sum_{k=1}^{K} ||{p_k} - proj( V_k, T_{cam}, R_{cam}, {In}_{cam})||_{2}\,
\end{equation}

\begin{equation} \label{eq: triplane}
{V_k} = triplane[v_k, u_k,:] 
\end{equation}
where $proj$ is an projection operation, $T_{cam}$ and $R_{cam}$ are translation and rotation of camera respectively, and ${In}_{cam}$ is fixed camera intrinsics. $u_k$ and $v_k$ are the uv-coordinate of vertex $V_k$.

The total variation loss is introduced to encourage local smoothness on mesh:
\begin{equation} \label{eq: tv_loss}
\mathcal{L}_{tv} = TV(triplane) 
\end{equation}

We additionally add symmetric loss on triplane to encourage facial symmetry for aesthetics consideration:
\begin{equation} \label{eq: sym_loss}
\begin{split}
\mathcal{L}_{sym} = ||triplane[...,0] + flip(triplane[...,0])) ||_{2} \\+ ||triplane[...,1] - flip(triplane[...,1])) ||_{2} \\+ ||triplane[...,2] - flip(triplane[...,2])) ||_{2}
\end{split}
\end{equation}

The total loss function on geometry generation are given by:
\begin{equation} \label{eq: total_loss}
\mathcal{L} =  \mathcal{L}_{lmk} + \lambda_{tv} \mathcal{L}_{tv} +\lambda_{sym}  \mathcal{L}_{sym}
\end{equation}

\begin{figure}[t]
  \centering
  \resizebox{0.98\linewidth}{!}{
\includegraphics{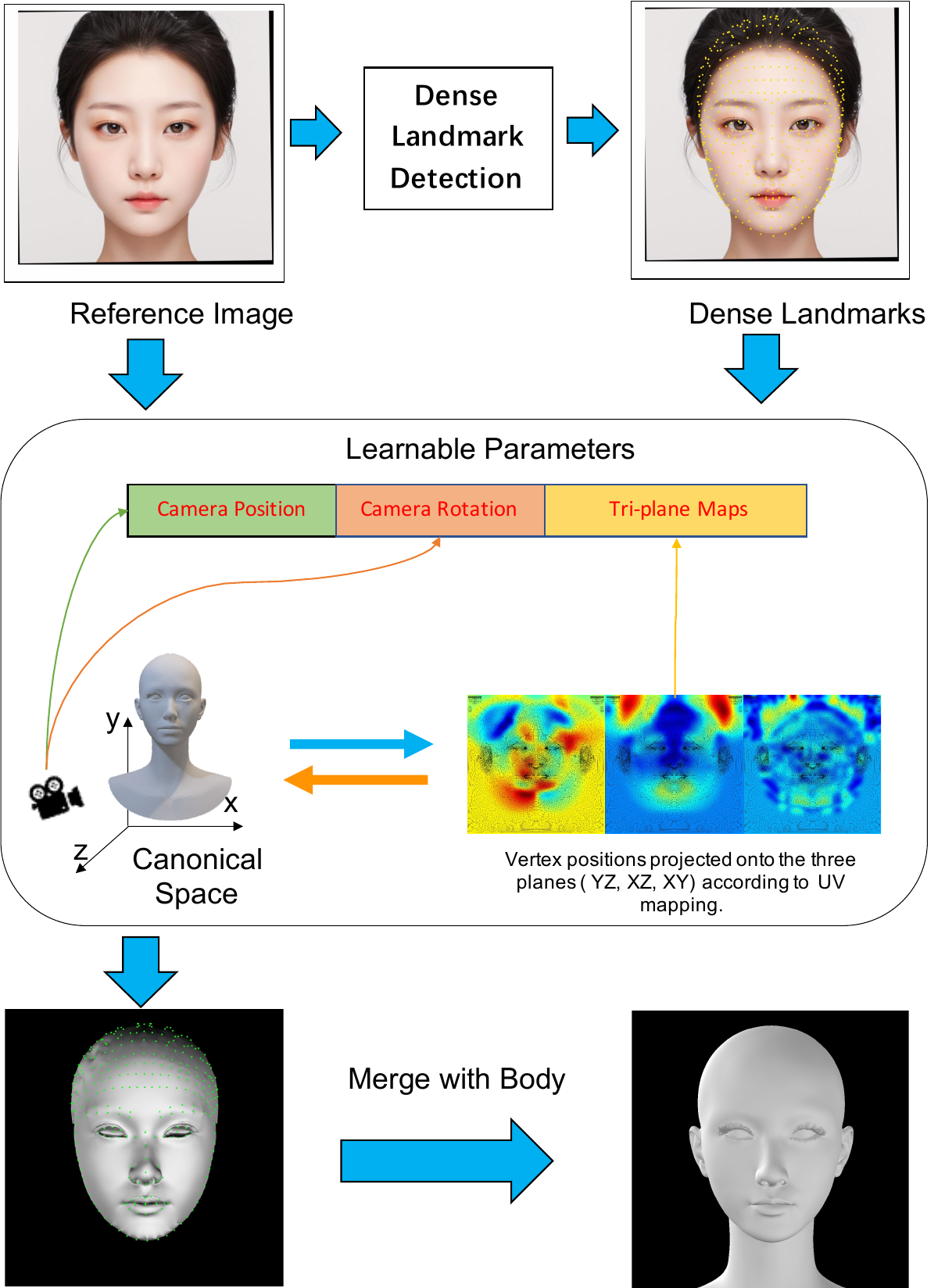} }
  % \vspace{-10pt}
  \caption{In the process of geometry generation, we optimize the camera parameters and triplane maps under the guidance of dense facial landmarks and the reference image. The position of each vertex is encoded into the triplane maps based on its corresponding UV coordinates.}
  \label{fig: geometry}
  % \vspace{-15pt}
\end{figure} 

\subsection{Texture Generation}

\subsubsection{Texture Extraction}
\begin{figure}[t]
  \centering
  \resizebox{0.98\linewidth}{!}{
   \includegraphics{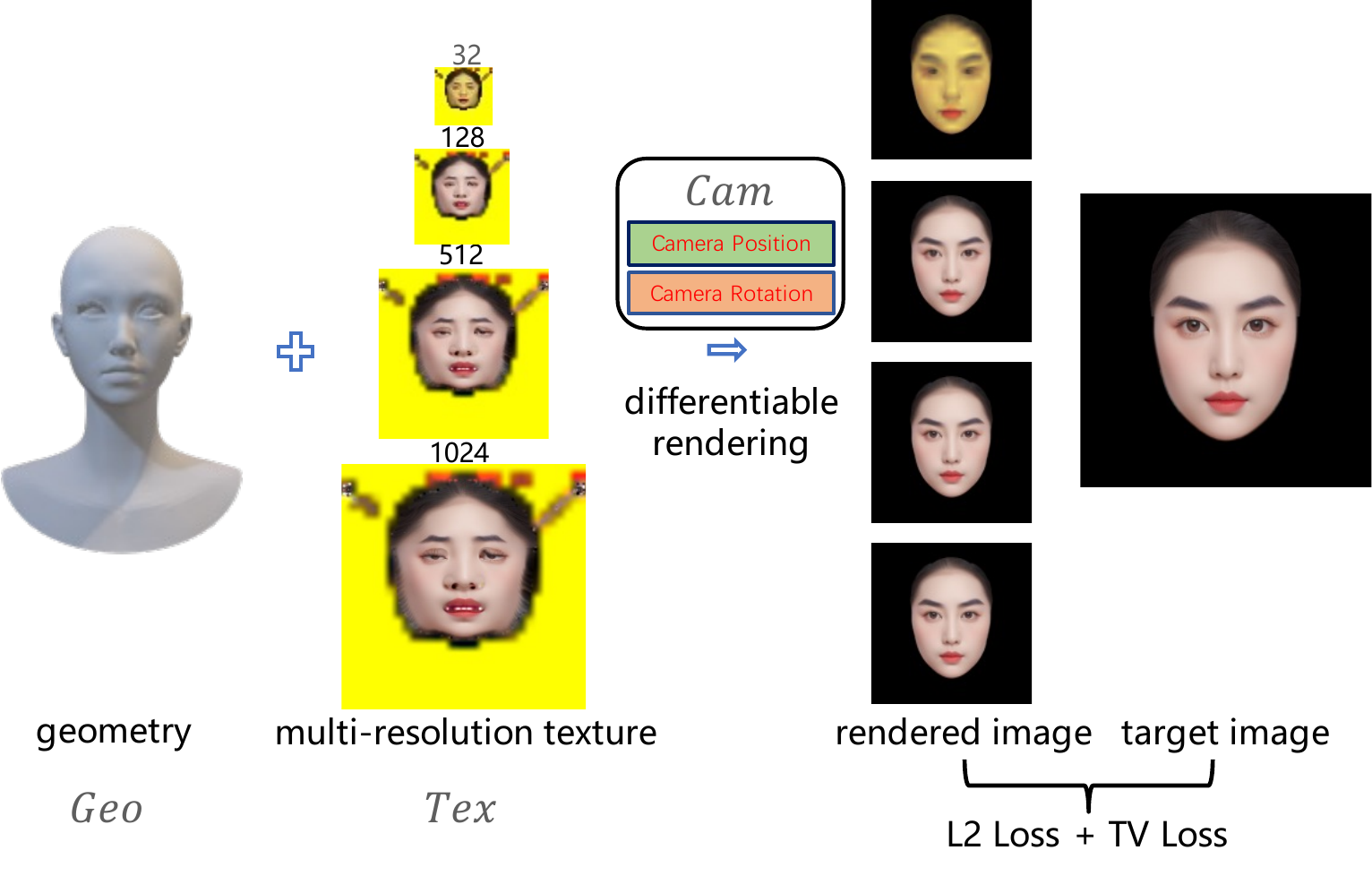} }
  % \vspace{-10pt}
  \caption{We employ a multi-resolution strategy, leveraging differentiable rendering methods, to obtain a detailed and coherent texture map. Our resolution hierarchy is defined as follows: [32, 128, 512, 1024].}
  \label{fig: gen_texture}
  \vspace{-15pt}
\end{figure}

After fitting the geometry to match the reference image, we employ differentiable rendering method to get the required texture image. The camera setting used here are kept consistent with geometry generation stage. Since there is not always a one-to-one correspondence between pixel positions and UV coordinates, we adopt a multi-resolution approach for texture generation by progressively generating texture from low to high resolution. As shown in Figure~\ref{fig: gen_texture}, $Geo$ and $Cam$ are the results of the geometry generation from the previous section, and $Tex$ denotes the target texture we aim to fit. The loss between the rendered image and the target image is calculated by the following equation:
\begin{equation} 
\label{eq: diff_render}
I_R = DR(Geo, Tex, {Cam})
\end{equation}
\begin{equation}\label{eq: tex_loss}
\mathcal{L}_{Tex} = ||I_R - I_T||_F + \alpha* TV(I_R)
\end{equation}
where, $DR$ denotes differentiable rendering,  $I_R$ is the rendered image, $I_T$ is the target image, $\alpha$ is a weight coefficient, here set to 0.01.

\subsubsection{Diffuse Albedo Estimation}
Using the texture obtained directly from differentiable rendering is not ideal, as it fails to fully separate the illumination and diffuse albedo components from the texture image. Consequently, when rendering under varied lighting conditions, the textures containing baked illumination may restult in noticeable unrealistic shadings and shadows. To tackle this issue, we introduce a neural delighting method to remove the undesired illumination from the texture image and get the render-ready diffuse albedo. It is noteworthy that our delighting algorithm works on texture images rather than portrait images. It is an intentional choice since texture images are free from occlusions and consistent across different poses and expressions of portraits, making the data acquisition and the algorithm learning more tractable.

\begin{figure}[t]
  \centering
  \resizebox{0.98\linewidth}{!}{
   \includegraphics{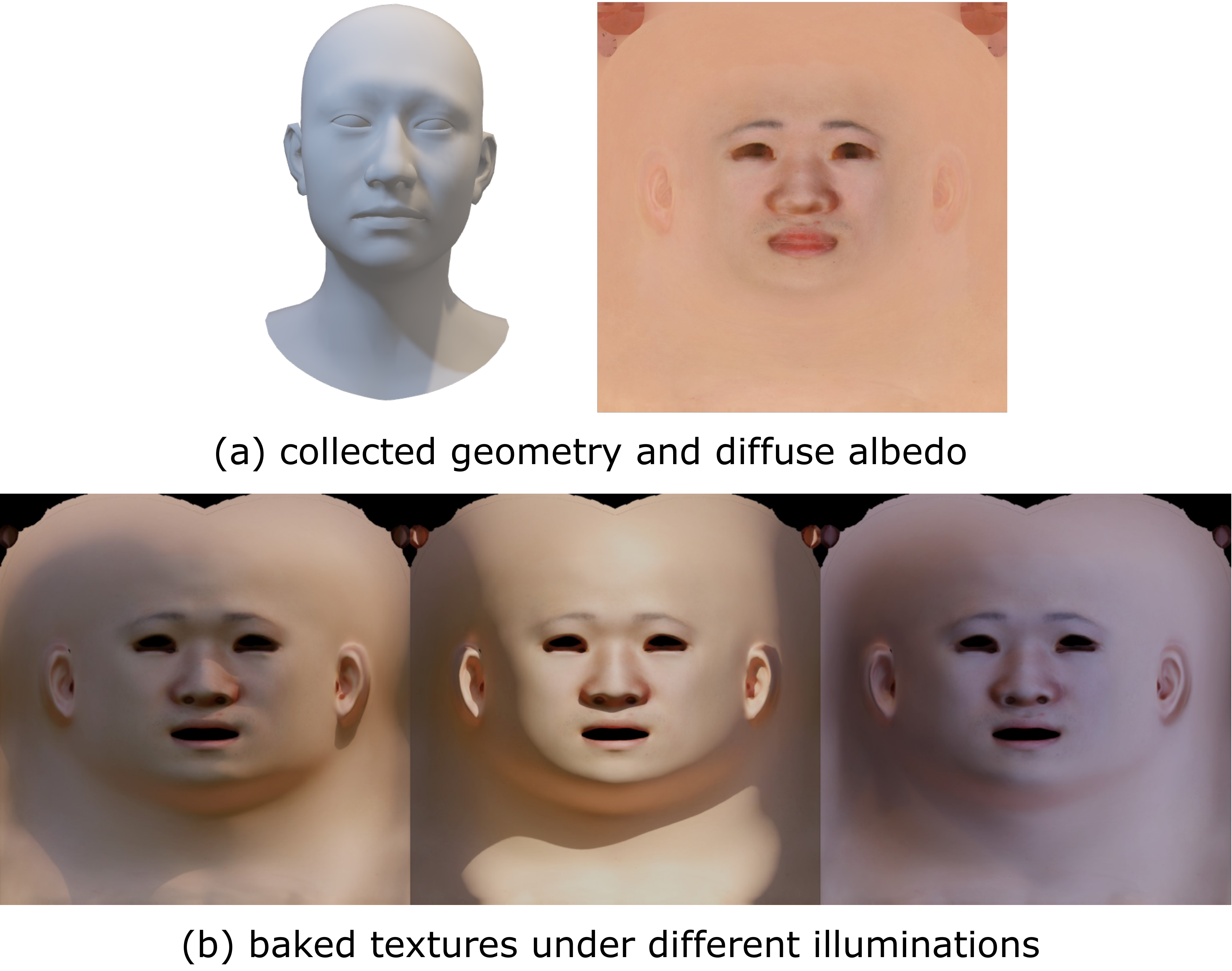} }
  % \vspace{-10pt}
  \caption{The data acquisition and processing pipeline.}
  \label{fig: gt_data}
  % \vspace{-15pt}
\end{figure}

\textbf{Ground Truth Data Collection.} 
As in \cite{Ghosh_Fyffe_Tunwattanapong_Busch_Yu_Debevec_2011}, we capture faces of 193 individuals (including 100 females and 93 males aged between 20-60 years old) under uniform illumination. By reconstructing the geometries, we obtain the unwarpped high resolution diffuse albedo, as in Figure~\ref{fig: gt_data} (a).

\textbf{Training Data Generation.} The textures under varying illumination are synthesized by baking lights into the ground truth diffuse albedo. In order to cover the wide range of natural lighting conditions, we bake 100 high dynamic range (HDR) lights (including indoor/outdoor, day/ night scenarios) for each ground truth data. To improve the data diversity and avoid overfitting, the skin colors of the ground truth diffuse albedos are augmented according to the Individual Typology Angle (ITA)\cite{CHARDON_CRETOIS_HOURSEAU_1991}. Figure ~\ref{fig: gt_data} (b) illustrates the baked textures in different illuminations.

\textbf{Delighting Network.} Without losing generality, we formulate the texture delighting problem as an image-to-image translation problem. Specifically, we employ the coarse-to-fine pix2pixHD network\cite{Wang_Liu_Zhu_Tao_Kautz_Catanzaro_2018}, which takes the synthesized illuminated textures as input and generates visually appealing high-resolution diffuse albedos. As in \cite{Wang_Liu_Zhu_Tao_Kautz_Catanzaro_2018}, the loss function is defined as a weighted combination of GAN loss and VGG feature matching loss. We train the network at the resolution of 1024 using the default parameters.

\subsubsection{Texture correction and completion}
\begin{figure}[t]
  \centering
  \resizebox{0.95\linewidth}{!}{
   \includegraphics{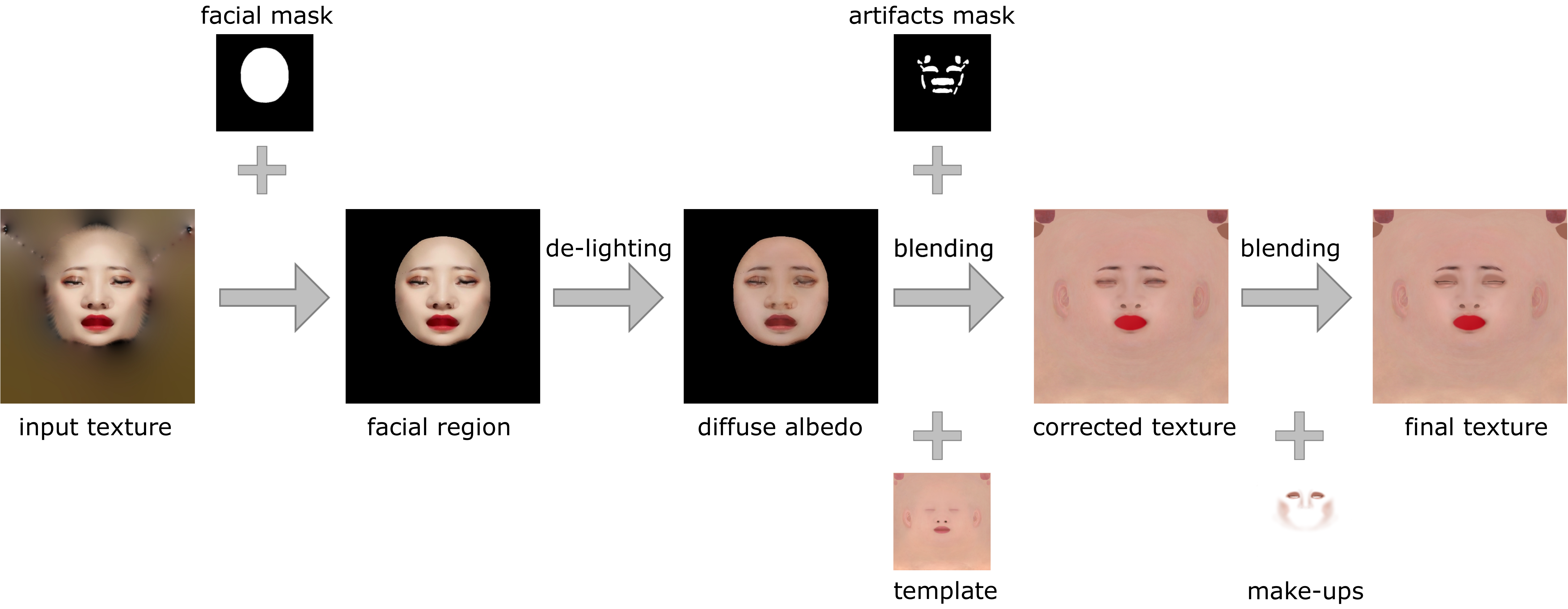} }
  % \vspace{-10pt}
  \caption{The proposed texture correction and completion pipeline.}
  \label{fig: texture_complete}
  \vspace{-15pt}
\end{figure}

After delighting, the generated diffuse albedo may still has artifacts in the vicinity of eyes, mouth and nostril regions. This is a result of the inherent limitations of the single frontal portrait image, which only provides limited textural information about a face. When mapped onto a 3D face geometry, the imperfect diffuse albedo introduces misaligned semantic features near eyes, mouth and nostril regions, leading to aesthetically unpleasant results. To address this issue, we utilize an off-the-shelf face parsing algorithm\cite{Yu_Wang_Peng_Gao_Yu_Sang_2018} to extract masks of these error-prone regions, which are then carefully dilated and merged with a template diffuse albedo using the Poisson method\cite{Pérez_Gangnet_Blake_2003}. Additionally, We transfer the colors of the mouth and eyebrows from the portrait image to maintain the facial features. Finally, the facial region is Poisson blended with the template diffuse albedo to obtain the textures of the ears and neck. We also add make-ups around eyes and cheeks to improve the aesthetics. Figure \ref{fig: texture_complete} demonstrates the proposed texture correction and completion pipeline.

\begin{figure}[t]
  \centering
  \resizebox{0.98\linewidth}{!}{
   \includegraphics{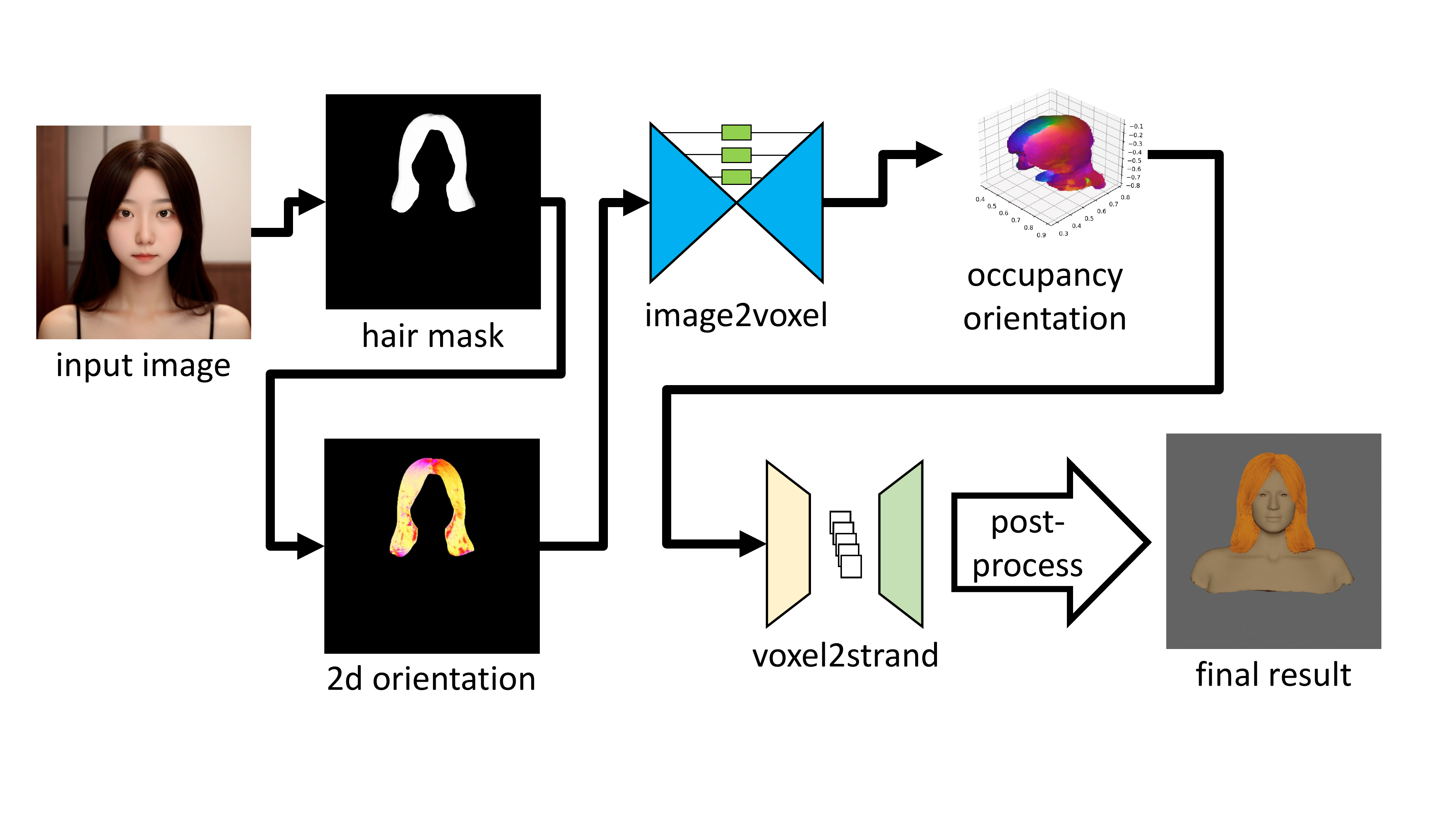} }
   \vspace{-10pt}
  \caption{Hair generation pipeline.}
  \label{fig: hair}
  \vspace{-15pt}
\end{figure}

\subsection{Hair Generation}
To produce a high-fidelity 3D avatar, we render hair as individual strands rather than as meshes. We firstly synthesis various hairstyle images via SD models,  and then conduct 3D strand-based hair reconstruction from these 2D images. We incorporate SOTA research such as NeuralHDHair~\cite{Wu_2022_CVPR}, and train our model using the USC-HairSalon~\cite{hu2015single} dataset. The hair reconstruction process consists of two main stages. Initially, the occupancy and orientation fields of the hair in 3D space are estimated based on the input 2D image. Subsequently, leveraging the discrete voxel-level data generated in the first phase, geometric descriptions of tens of thousands of hair strands are generated. The complete hairstyle generation procedure also involves pre-processing operations such as face alignment, hair mask segmentation, and 2D orientation recognition. Additionally, post-processing operations are applied to enhance the initial generated hair results, including the removal of unreasonable strands, fitting the hair to the target head mesh, and performing geometric deformations to achieve high-quality reconstruction results. The overall pipeline is illustrated in the Figure~\ref{fig: hair}. 

Considering that real-time hairstyle generation is time-consuming, we opt to generate diverse hairstyle assets offline. These generated hair assets, along with existing metahuman hairs, are labeled with descriptive attributes such as hairstyle type, length, and degree of crimp. This attribute labeling enables efficient matching processes.

% The USC-HairSalon dataset enables the model to have good ability to generate 3D strand-level hair from a single 2D image. However, for specific hair styles, the improvement of reconstruction capabilities depends on the introduction of more high-quality datasets.

\begin{figure}[t]
  \centering
  \resizebox{0.98\linewidth}{!}{
   \includegraphics{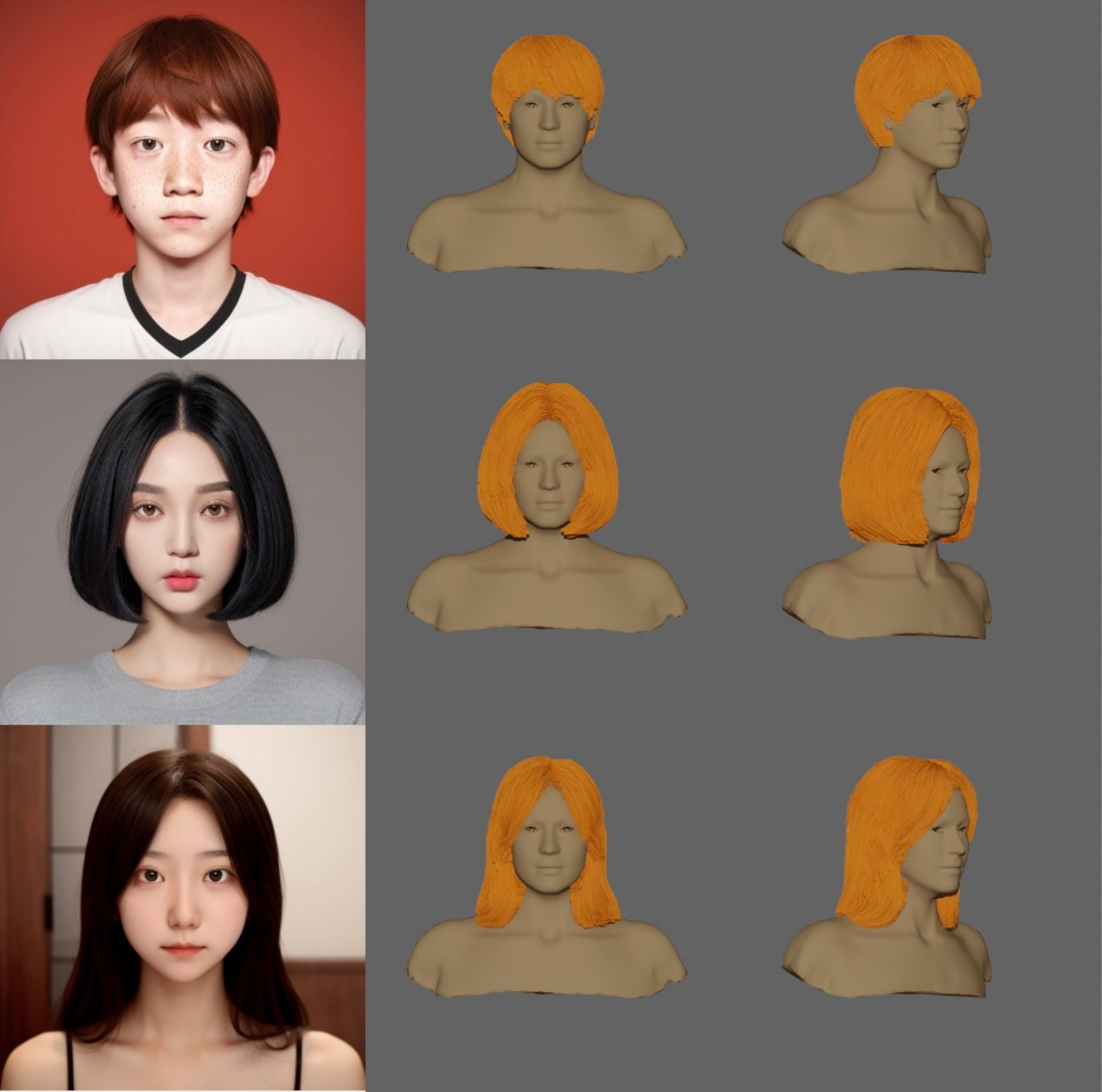} }
   \vspace{-10pt}
  \caption{Strand-based hair generation is guided by hairstyle images that are generated using SD models.}
  \label{fig: hair_generation}
  \vspace{-15pt}
\end{figure}

\subsection{Assets Matching}
To construct a fully-realized 3D character, we must integrate the generated head, hair, body, garments, and some accessories together. Each pre-produced asset is labeled with textual annotations, either manually or through annotation algorithm. To select the most suitable asset that matches the input prompt, we employ CLIP's text encoder~\cite{Radford2021LearningTV} to compute the cosine similarity between the features of the input prompt and the labels of the assets, the asset with the highest similarity is then selected.

\section{Results}
\label{sec:Results}

\begin{figure*}[ht]
  \centering
  \resizebox{0.98\linewidth}{!}{
   \includegraphics{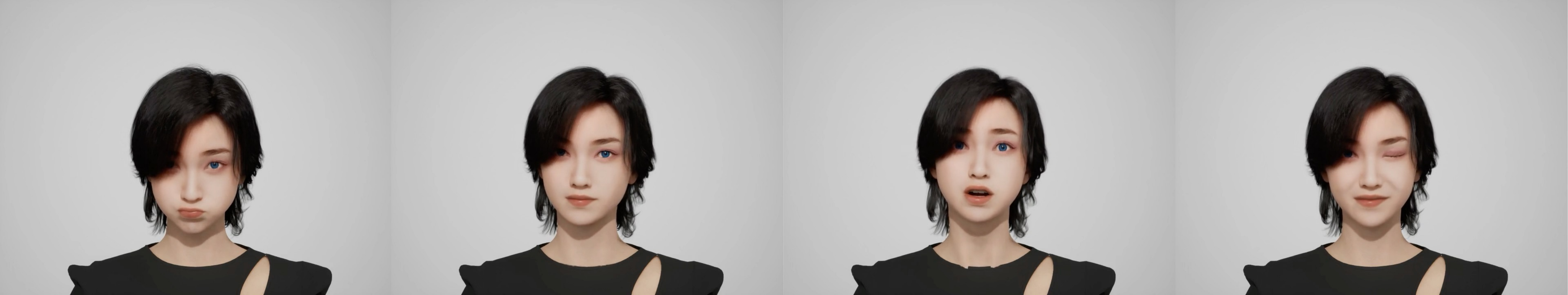} }
  \caption{Dynamic expression by employing facial rig control system~\cite{RigLogic}.}
  \label{fig: expression_showcases}
  \vspace{-15pt}
\end{figure*}

We present the visual results of the generated 3D avatars in Figure~\ref{fig: generation_showcases}, accompanied by respective input prompts listed below each avatar. In Figure~\ref{fig: expression_showcases}, we showcase the expressive animations achieved through facial rig control. These showcases were rendered using the Unreal Engine.

\section{Future Work}
\label{sec:Future}
Our current version focuses on generating visually appealing 3D avatars of Asian ethnicity, as our selected SD model is primarily trained on Asian facial images. In the future, we will try to expand support for different ethnicities and styles.
It is worth noting that our de-lighting datasets consist of clean face textures only, non-natural facial patterns like scribbles or stickers may be weakened in the generated avatars. Currently, our garments and body parts are pre-produced and matched based on textual similarity. However, we are actively working on developing cloth, expression, and motion generation techniques driven by text prompts.

{\small
\bibliographystyle{ieeenat_fullname}
\bibliography{11_references}
}

%\ifarxiv \clearpage \appendix \input{12_appendix} \fi

\end{document}

%% TITLE
\title{\paperTitle}
\author{\authorBlock}
\maketitlesupplementary

\appendix
\section{Appendix Section}
\label{sec:appendix_section}
Supplementary material goes here.

{\small
\bibliographystyle{ieeenat_fullname}
\bibliography{11_references}
}